\documentstyle[11pt,parsetree]{tmi}

\input{psfig}

\newtheorem{definition}{Definition}

\newcommand{\tab}{\hspace*{2em}}

\newcommand{\avmplus}[1]{{\setlength{\arraycolsep}{0.4mm}       
                       \renewcommand{\arraystretch}{0.7}
                       \left[                   
                       \begin{array}{l}
                       \\[-2mm] #1 \\[-2mm] \\
                       \end{array}              
                       \right]
                    }}

\newcommand{\attval}[2]{{\mbox{\scriptsize {\sc #1}}\! =\!\! {{#2}}}}
\newcommand{\attvaltyp}[2]{{\mbox{\scriptsize{\sc #1}}\! =\! {\myvaluebold{#2}}}}

\newcommand{\myvaluebold}[1]{{\mbox{\scriptsize {\bf #1}}}}

\newcommand{\ind}[1]{{\setlength{\fboxsep}{0.5mm} \fbox{{\scriptsize #1}} \!}}

\begin{document}
\title{Determining Internal and External Indices for Chart Generation}

\author{Arturo Trujillo}
\institute{CCL, UMIST, PO Box 88\\
Manchester M60 1QD, England\\
{\tt iat@ccl.umist.ac.uk}}

\maketitle

\begin{abstract}
  This paper presents a compilation procedure which determines
  internal and external indices for signs in a unification based
  grammar to be used in improving the computational efficiency of
  lexicalist chart generation. The procedure takes as input a grammar
  and a set of feature paths indicating the position of semantic
  indices in a sign, and calculates the fixed-point of a set of
  equations derived from the grammar. The result is a set of
  independent constraints stating which indices in a sign can be bound
  to other signs within a complete sentence. Based on these
  constraints, two tests are formulated which reduce the search
  space during generation.
\end{abstract}

\section{Introduction}

One problem with the classical transfer approach to machine
translation (MT) is that it involves complex transformations of
syntactic and semantic structures from the source to the target
language. These transformations can have intricate interactions with
each other, making transfer modules difficult to reverse, debug and
maintain. They can also make monolingual components more heavily
dependent on the language pair at hand.
\hyphenation{and}
\hyphenation{representations}
Much of the complexity in transfer stems from the recursive nature of
the syntactic and semantic frameworks normally used.  However, recent
work in formal semantics has found it expedient to minimise the
recursive structure of semantic representations to efficiently encode
certain types of ambiguity \cite{reyle95}. Naturally, flat semantics
mitigate many structural differences between natural languages and
their application to MT readily follows (see \cite{copestakeetal95b}).
Unfortunately, simplicity in the transfer component comes at the cost
of generation complexity for such representations since their lack of
structure increases the non-determinism of most generation algorithms,
just as lexical-only transfer increases the complexity of bag
generation in Shake-and-Bake MT \cite{whitelock94}. For this reason,
several researchers have investigated the efficiency of generators
whose input has a flat structure, be this in the form of lists of
semantic predicates or of lexical elements. Such generators, of the
chart, bag and lexicalist varieties, differ in many ways but the
source of their complexity is the same: the search space grows
factorially on the size of the input for many algorithms, since they
are based on a modified chart parser which essentially attempts all
permutations of the input. This is the issue addressed by the paper,
taking chart generation as an instance of the problem.

\section{Chart Generation}

A chart generator \cite{kay96} takes as input a flat semantic
representation and, using a chart data structure, outputs the string
corresponding to it. The unordered character of the semantic input
permits such generators to be viewed as parsers for languages with
completely free word order: an active edge combines with an inactive
edge only if the two edges have no semantic predicates in common; no
other restrictions apply. This regime leads to the combinatorial
explosion mentioned above.

\subsection{Example}

Consider the following flat semantic representation corresponding to
the string {\em John ran fast}:
\begin{quote}
  r : run(r), past(r), fast(r), arg1(r,j), name(j,John)
\end{quote}
Here, $r$ is the distinguished index for the expression. These predicates will
unify with the semantic component of suitably defined lexical entries resulting
in the agenda entries shown below:

\hspace*{4em}  \begin{tabular}{lll} \hline
    Word & Cat & Semantics \\ \hline \hline John & np(j) & j :
    name(j,John) \\ ran & vp(r,j) & r : run(r), arg1(r,j), past(r) \\ 
    fast & adv(r) & r : fast(r) \\ \hline
  \end{tabular}

Items are then moved into the chart and their interactions considered. Moving
{\em John} results in no interactions, since the chart is empty. Moving
{\em ran} results in {\em John ran} assuming the rule:
\begin{quote}
  s(x) $\rightarrow$ np(x), vp(x,y)
\end{quote}
This is a complete sentence, but it does not subsume all the semantic
material from the input; it therefore remains in the chart but cannot
constitute an output sentence.  Next, {\em fast} is moved, adding
in {\em ran fast} to the agenda and then to the chart, at
which point {\em John ran fast} is built. Generation thus terminates.
One of the main sources of inefficiency in chart generation is that a
multitude of edges are constructed which either do not subsume the
entire semantics of the input or which can never be part of the
solution because they omit semantic material which only they could
have subsumed. In the example, {\em John runs} is one such edge. The
problem is that these edges, if left in the chart, will interact with
other edges to form yet further edges which can never be part of the
final result, but which cause the search space to explode.

\subsection{Internal and External Indices}
\label{inn-out-sec}

To overcome this inefficiency, it is necessary to discard edges which
would make it impossible to incorporate all the input into an output
sentence.  Achieving this involves exploiting the fact that after an
edge is constructed, only certain indices in its semantic predicates
are accessible by other rules in the grammar. For example, treatments
of English VPs (e.g. {\em chased the cat}) typically disallow
modification of the object NP once the VP has been analysed; thus if
{\em cat} received index {\em c}, it is not possible to bind into this
index. Intuitively, this means that modifying the VP cannot lead to
modification of the object NP.
Following Kay, indices not available outside a category (i.e. outside
an inactive edge) are called {\em internal} indices, while those which
are accessible are called {\em external} indices.  When an inactive
edge is constructed, all indices in predicates not subsumed by the
edge must be i) different from the indices the edge subsumes, or ii)
be external to it.  This ensures that inactive edges subsume all
predicates indexed by their internal indices.
The objective of this paper is to present a general algorithm for
determining which indices are internal and external to a category
without requiring the explicit identification of such information by
the grammar writer.

\section{Overview of the Algorithm}

Ideally, internal indices should be determined directly from the rules
of a grammar. However, different grammar writers adopt different index
binding strategies, making such identification by automatic on-line
inspection of rules very difficult at best. The algorithm proposed
here therefore automatically extracts information from a grammar
off-line and uses it to determine whether an index is internal or
external to a category. Based on this information, it is possible to
identify those edges which are incomplete with respect to the input
and which may consequently be eliminated from further consideration.
The algorithm has been implemented and tested on a lexicalist
generator operating on a small unification-based grammar; a
description of the test and further discussion of the issues involved
is given in \cite{trujilloetal96}.
The algorithm takes as input a unification-based phrase structure (PS)
grammar and a set of paths and outputs a set of constraints on pairs
of signs indicating which indices in the two signs can be bound for
some possible derivation tree. Principal among the techniques used are
those for predictive parser compilation \cite{ahoetal86} adapted to
unification based grammars \cite{trujillo94}. In addition, following
standard practice in data flow analysis \cite{kennedy81}, a data
structure is maintained tracing how variables (or in this case,
indices) are modified (or in this case, bound) in a valid
derivation.

\subsection{Inner and Outer Domains}

Two main phases, themselves analogous to the calculation of FIRST and
FOLLOW sets for predictive parsers, constitute the bulk of the
algorithm. The first phase determines the indices at the root of a
tree which are bound to items at the leaves; this phase will be called
the calculation of {\em inner domains}.  The second phase uses inner
domains to calculate {\em outer domains}, which indicate the indices
in a sign which are bound to the indices 
of signs outside the sign's subtree.
Thus, inner domains express the relationship between phrases with
related semantic material within subtrees for which they are roots,
while outer domains express the relationship between a phrase and
outside phrases with which the phrase shares semantic material.
Calculating both inner and outer domains requires the computation of
the fixed-point of a set of equations derived from the grammar. The
fixed-point of a function is the value of {\bf X} which satisfies \\ 
\tab f({\bf X}) = {\bf X}

\subsection{Grammar}

We adopt the following definition of a unification grammar:
\begin{definition}
A grammar is a tuple (N,T,P,S), where P is a set of productions
$\alpha \Rightarrow \beta$, where $\alpha$ is a sign, $\beta$ is a
list of signs, N is the set of all $\alpha$, T is the set of all signs
appearing in $\beta$ such that they unify with lexical
entries, and S is the start sign.
\end{definition}
The grammar must generate sequences of coherent predicates (i.e. the
graph with arcs for predicates sharing indices is connected).

\subsection{The Triple Data Structure}

A basic data structure in the algorithm will be triples of the form
{\em (Left Sign, Right Sign, Bindings)}, where {\em Bindings} is a set
of pairs consisting of a path in {\em Left Sign} and a path in {\em
  Right Sign} such that the values at the end of each path are assumed
to be token identical; {\em Left Sign} and {\em Right Sign} are
phrasal or lexical signs. The following triple for example represents 
part of the inner domain of an NP:
\begin{quote}
(1)  (NP[sem:arg1:X],Det[sem:arg1:Y], \{$<$sem:arg1,sem:arg1$>$\})
\end{quote}
It indicates that in a complete parse, it is possible that index X be
bound to index Y for these two signs.  Triples of this form are used
uniformly throughout to encode inner and outer domains and to
represent the functions and equations for which a fixed-point needs to
be found.
The algorithm proper consists of three main components: general operations,
inner domain compilation and outer domain compilation.

\section{General Operations}

\subsection{Fixed-Point Iteration}

This is the key function in the compilation process and it is used to
solve systems of equations derived from the grammar. Each such system
can be interpreted as a vector function \cite{rayward-smith83} with
one side of the equations used to calculate a value which is then
assigned to the other side.
The fixed-point algorithm 
takes a function and
an initial argument value (both expressed as sets) and returns, also
as a set, the result of iteratively applying the function to
successive values:\\
{\em Fixed-point(Function, Argument) $\rightarrow$ Set of triples\\
\tab Result := \{\} \\
\tab A-new := Argument \\
\tab Repeat \\
\tab \tab Temp := Function X A-new \\
\tab \tab Result$'$ := Result $\cup_{\leq}$ Temp \\
\tab \tab A-new := Result$' -$ Result (i.e. set difference) \\
\tab \tab Result := Result' \\
\tab Until A-new := \{\} \\
\tab Return Result }

Two (overloaded) operators are used in this algorithm:
\begin{enumerate}
\item The crossproduct operator, {\em X}, takes two sets, {\em A} and {\em B} and
constructs the set $\{ a \times b \mid a \in A, b \in B \}$, where $\times$ is 
a type dependent operation, defined as follows:
\begin{itemize}
\item If $a$ and $b$ are triples of the form {\em (La,Ra,Ba)}
      and {\em (Lb,Rb,Bb)}, then\\
      $a \times b = (La,Rb, Ba X Bb)$ if  $Ra \sqcap Lb$ (i.e. they unify) and $Ba 
      \times Bb \neq \{\}$ 
\item If $a$ and $b$ are pairs of paths of the form $<Lpa,Rpa>$ and $<Lpb,Rpb>$, 
      then\\
      $a \times b = <Lpa,Rpb>$ if $Rpa$ is equal to $Lpb$.
\end{itemize}
For example, the following operations indicates that if a PP is in
the outer domain of a VP, so is a preposition:
\begin{quote}
  \{ (VP[sem:arg1:W],PP[sem:arg1:X],\{$<$sem:arg1,sem:arg1$>$\}) \}\\
   X  \{ (PP[sem:arg1:Y],P[sem:arg1:Z],\{$<$sem:arg1,sem:arg1$>$\}) \} \\
   =   \{ (VP[sem:arg1:W],P[sem:arg1:X],\{$<$sem:arg1,sem:arg1$>$\}) \}\\
\end{quote}

\item The subsume-union operator, $\cup_{\leq}$, takes two sets, $A$
  and $B$ and calculates the set $\bigcup\{ a \sqcup_t b \mid 
      a \in A, b \in B\}$, where $\sqcup_t$ is a type dependent generalisation 
      operator, defined as follows:
  \begin{itemize}
  \item If $a$ and $b$ are triples of the form {\em (La,Ra,Ba)}
      and {\em (Lb,Rb,Bb)} then $a \sqcup_t b$ is
    \begin{itemize}
    \item {\em \{ (La,Ra,Bab) \}}, where $Bab =  Ba \cup_{\leq} Bb$,
      if $Lb \sqsubseteq La$ (i.e. $La$ subsumes $Lb$) and $Rb \sqsubseteq Ra$.
    \item {\em \{ (Lb,Rb,Bab) \}}, where $Bab =  Ba \cup_{\leq} Bb$,
      if $La \sqsubseteq Lb$ and $Ra \sqsubseteq Rb$.
    \item {\em \{ (La,Ra,Bab), (Lb,Rb,Bab) \}} otherwise.
    \end{itemize}
  \item If $a$ and $b$ are pairs of paths of the form $<Lpa,Rpa>$ and $<Lpb,Rpb>$,
    then $a \sqcup_t b$ is
    \begin{itemize}
    \item $\{<Lpa,Rpa>\}$ if $Lpa$ is a prefix of path $Lpb$ and $Rpa$ is a prefix
     of path $Rpb$.
    \item $\{<Lpb,Rpb>\}$ if $Lpb$ is a prefix of path $Lpa$ and $Rpb$ is a prefix
     of path $Rpa$.
    \item $\{<Lpa,Rpa>, <Lpb,Rpb>\}$ otherwise.
    \end{itemize}
  \end{itemize}
For example, the fact that prepositions can modify the event of a VP and 
also its subject, leads to the following union:
\begin{quote}
  \{ (VP[sem:[arg1:U,arg2:V]],P[sem:arg1:W],\{$<$sem:arg1,sem:arg1$>$\}) \}\\
  $\cup_{\leq}$ \{ (VP[sem:[arg1:X,arg2:Y]],P[sem:arg1:Z],\{$<$sem:arg2,sem:arg1$>$\}) \}
\end{quote}
   = \{(VP[sem:[arg1:U,arg2:V]],P[sem:arg1:W],\{$<$sem:arg2,sem:arg1$>$,$<$sem:arg1,sem:arg1$>$\})\}

\end{enumerate}

\subsection{Shared Indices}

When constructing the equations for inner and outer domains from
grammar rules, the index paths shared between categories in a rule need to be
identified. A dedicated function achieves this:\\
{\em Shared-Indices(Sign1,Sign2,Paths) $\rightarrow$ \\
\tab \{ $<$p1,p2$> \mid$ Sign1:p1 is token identical with Sign2:p2, and p1 and p2\\
\tab are both prefixes of elements in Paths\} }\\
For example:\\
 {\em Shared-Indices}(NP[sem:arg1:X],Det[sem:arg1:X],\{sem:arg1, sem:arg2\}) \\
= \{$<$sem:arg1,sem:arg1$>$\}

\section{Compiling Inner Domain}

Computing inner domains proceeds by finding the fixed-point of inner
domain equations derived from the grammar, represented as triples.
For instance, triple 1) above would be one of the equations for the rule:
\begin{quote}
$\avmplus{
\attvaltyp{cat}{NP}\\
\attval{sem}{\ind{0}}}
\Longrightarrow
\avmplus{
\attvaltyp{cat}{Det}\\
\attval{sem:arg1}{\ind{1}}}
\hspace{1em}
\avmplus{
\attvaltyp{cat}{N1}\\
\attval{sem}{\ind{0}}\avmplus{
                        \attval{arg1}{\ind{1}}}}$
\end{quote}
The triple would therefore indicate that the inner domain of an {\em
  NP} includes the inner domain of a {\em Det}. In the case of inner
domains, the equations can also be interpreted as the initial value to
be fed to {\em Fixed-point}; that is, triple 1) can also be interpreted as saying
that the inner domain of {\em NP} includes {\em Det}.
Inner domain equations are built as follows:
\begin{quote}
{\em Inner-Equations(Grammar) $\rightarrow$ Set of triples\\
\tab Inner-Eq := \{\}\\
\tab For each rule $A \rightarrow B_1 ... B_n \in$ Grammar\\
\tab \tab Inner-Eq := Inner-Eq 
$\cup {\displaystyle \bigcup_{k=1..n}}\{(A,B_k,$Shared-Indices$(A,B_k,Paths))\}$\\
\tab Return Inner-Eq }
\end{quote}
Here, {\em Paths} is a theory specific set of index paths.
The set of inner domains can now be defined as:
\begin{quote}
{\em Fixed-point(Inner-Equations(Grammar),Inner-Equations(Grammar))}
\end{quote}

\section{Compiling Outer Domains}

Outer domains encode the bindings that may exist between a sign and any signs
external to it in a valid derivation. The tree:\\
{\scriptsize
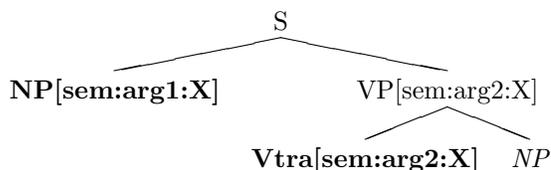
\begin{parsetree}
(.S. `{\bf NP[sem:arg1:X]}'
     (.VP[sem:arg2:X]. `{\bf Vtra[sem:arg2:X]}'
                       `NP')) 
\end{parsetree}}

\noindent
would give rise to the following triple in the outer domain set:
\begin{quote}
(2)  (NP[sem:arg1:X],Vtra[sem:arg2:Y], \{$<$sem:arg1,sem:arg2$>$\})  
\end{quote}
This states that Vtra is in the outer domain of NP because the X in
the NP can be bound to Y in Vtra. Outer domain equations are calculated
as follows:
\begin{quote}
{\em Outer-Equations(Grammar) $\rightarrow$ Set of triples\\
\tab Outer-Eq := \{\}\\
\tab For each rule $A \rightarrow B_1 ... B_n \in$ Grammar\\
\tab \tab Outer-Eq := Outer-Eq $\cup 
               {\displaystyle \bigcup_{k=1..n}}\{(B_k,A,$Shared-Indices$(B_k,A,Paths))\}$\\
\tab Return Outer-Eq }
\end{quote}
Each triple in these equations states that the outer domain of a sign
in a rule is the outer domain of its mother.
The initial value for outer domains can be calculated from the grammar and the
set of inner domains:
\begin{quote}
{\em Initialise-Outer-Domains(Grammar) $\rightarrow$ Set of triples\\
\tab Outer-Dom := \{\}\\
\tab For each rule $A \rightarrow B_1 ... B_n \in$ Grammar\\
\tab \tab Outer-Dom := Outer-Dom $\cup {\displaystyle \bigcup_{1\leq j,k\leq n, j \neq k}}$\{ (B$_j$,B$_k$, 
           Shared-Indices(B$_j$,B$_k$) \}\\
\tab Outer-Dom := \\
\tab \tab Outer-Dom X Fixed-point(Inner-Equations(Grammar),Inner-Equations(Grammar)) \\
\tab Return Outer-Dom }
\end{quote}
I.e. the outer domain of a category is the inner domain of all its
sisters within a rule. Once initialised, the outer domains can be
computed with:
\begin{quote}
  {\em Fixed-point(Outer-Equations(Grammar),Initialise-Outer-Domains(Grammar)) }
\end{quote}

\section{Using Outer Domains}

Once calculated, outer domains can be used in at least two ways for
chart generation. Firstly, they can be used to determine the internal
indices of an edge and thus identify predicates which may have been
`left out' (see Section \ref{inn-out-sec}).  To compute the internal
indices for each inactive edge, the edge's external indices are
subtracted from all the indices subsumed by it.  External indices can
be determined via outer domain triples through the use of the left
sign and the bindings set.  The following predicate implements the
relevant test; it assumes that as edges are built, a record of the
indices found amongst its predicates is kept and accumulated as more
complex edges are built. The function returns true if an edge includes
all the predicates indexed by its internal indices.
\begin{quote}
{\em Internal-Validation(Inact-Edge,Remaining-Preds,Outer-Domain) $\rightarrow$ Boolean\\
\tab External-Indices := \{\}\\
\tab For each triple (Inact-Edge, \_ ,Binds) $\in$ Outer-Domain\\
\tab \tab For each pair $<p,\_> \in$ Binds \\
\tab \tab \tab Add the index at the end of Inact-Edge:p to External-Indices \\
\tab Internal-Indices := Inactive-Edge.indices $-$ External-Indices\\
\tab Return false if there is a predicate in Remaining-Preds indexed by \\
\tab an element of Internal-Indices\\
\tab Return true otherwise}
\end{quote}
For example, given the logical form for {\em the dog saw the white cat}:
\begin{quote}
(3) s : def(d), dog(d), see(s,d,c), past(s), def(c), cat(c), white(c)
\end{quote}
the edge {\em saw the cat} would be discarded because there would be
no triple in which the third argument index in a VP is bound to an index
in any category.
Thus, index $c$ would be deemed internal to
the VP, and, since predicate {\em white(c)} includes this index, the
VP could not be part of an output sentence.
One disadvantage of this test is that it takes no account of the
category of the signs outside the inactive edge and therefore allows
too many unnecessary edges. Thus, while {\em saw the cat} is
eliminated, the edge {\em the cat} is still constructed because $c$
will be an external index to the NP.  
The second test is designed to exploit the information in the outer
domains to detect such edges. The following function returns true if
the external indices in an inactive edge can indeed be bound to
indices on external signs.  The main idea is that any lexical items
indexed by external indices must be licenced by at least one triple in
the outer domain set.
\begin{quote}
{\em External-Validation(Inact-Edge,Remaining-Preds,Outer-Domain) $\rightarrow$ Boolean\\
\tab For each LexSign $\in$ Remaining-Preds which includes an external index from\\
\tab Inact-Edge\\
\tab \tab If there is no triple (Inact-Edge,LexSign,Binds) $\in$ Outer-Domain for\\
\tab \tab which Inact-Edge:a = LexSign:b for at least one $<a,b> \in$ Binds for all \\
\tab \tab external indices in LexSign, then\\
\tab \tab \tab Terminate the loop and return False \\
\tab Return True}
\end{quote}
To disallow {\em the cat} when generating from (3), the outer domain
set is scanned for a triple with NP as its left sign, Adj as its right
sign and a pair of paths binding index $c$ in Adj to the index in the
NP; since no such triple would be present, Adj could not be
incorporated into the semantics of a sentence including this NP.
Therefore, the NP is discarded.
During generation, a conjunction of both predicates needs to be
satisfied before a newly constructed inactive edge can be added to the
chart:
\begin{quote}
Internal-Validation(Inact-Edge,Remaining-Preds,Outer-Domain) $\wedge$ \\
External-Validation(Inact-Edge,Remaining-Preds,Outer-Domain)
\end{quote}

\section{Evaluation and Conclusion}

\hyphenation{repecti-vely}
The algorithm was implemented in Sicstus Prolog and used to compile
the outer domains for a small unification-based grammar; the resulting
outer domains were tested on a lexicalist chart generator
\cite{trujilloetal96}. The grammar handles adjectival and
prepositional modification, both of which are common in real text.
Relative clauses and gapping in general are not handled fully since
they cause a larger number of indices to become external.  On a small
corpus of 10 sentences (average length = 9.8 words), use of internal
and external validation reduced average generation time and final
number of edges in the chart by 32\% and 27\% respectively. The best
improvements were for the sentence {\em the big brown dog with the
  fancy collar chased a little cat in the middle of the afternoon}
(221 secs to 39 secs; 852 edges to 272); the worst performance came
from {\em the man
  employed the woman} (29 secs to 30 secs; 50 edges to 50 edges).
While the algorithm is generally applicable to grammars with a strong
PS component, further work is required to extract inner and outer
domains from purely lexicalized grammars such as UCG or hybrids 
like HPSG. The fixed-point algorithm is general and by modifying the
equations constructed, it can be used for compiling parsing tables for
unification-based grammars \cite{trujillo94}.

\section*{Acknowledgements}

Thanks to two anonymous reviewers and to John Carroll, Nicolas Nicolov
and the staff and students at ITRI, University of Brighton, for useful
comments.

\bibliographystyle{fullname}

\bibliography{ref}

\begin{thebibliography}{}

\bibitem[\protect\citename{Aho {\em et al.}}1986]{ahoetal86}
Aho, A.~V., R.~Sethi, and J.~D. Ullman.
\newblock 1986.
\newblock {\em Compilers - Principles, Techniques, and Tools}.
\newblock Addison Wesley, Reading, MA.

\bibitem[\protect\citename{Copestake \bgroup et al.\egroup
  }1995]{copestakeetal95b}
Copestake, A., D.~Flickinger, R.~Malouf, S.~Riehemann, and I.~Sag.
\newblock 1995.
\newblock Translation using minimal recursion semantics.
\newblock In {\em Proceedings of the Sixth International Conference on
  Theoretical and Methodological Issues in Machine Translation}, pages 15--32,
  Leuven, Belgium.

\bibitem[\protect\citename{Kay}1996]{kay96}
Kay, M.
\newblock 1996.
\newblock Chart generation.
\newblock In {\em 34th Annual Meeting of the Association for
  Computational Linguistics}, pages 200--04, Santa Cruz, CA.

\bibitem[\protect\citename{Kennedy}1981]{kennedy81}
Kennedy, Ken.
\newblock 1981.
\newblock A survey of data flow analysis techniques.
In Muchnick, Steven~S. and Neil~D. Jones, editors.
\newblock {\em Program Flow Analysis: Theory and Applications}.
\newblock Software. Prentice-Hall, Englewood Cliffs, NJ, chapter~1, pages 5--54.

\bibitem[\protect\citename{Rayward-Smith}1983]{rayward-smith83}
Rayward-Smith, V.~J.
\newblock 1983.
\newblock {\em A First Course in Formal Language Theory}.
\newblock Computer Science Texts. Blackwell Scientific Publications, Oxford,
  UK.

\bibitem[\protect\citename{Reyle}1995]{reyle95}
Reyle, Uwe.
\newblock 1995.
\newblock On reasoning with ambiguities.
\newblock In {\em Seventh Conference of the European Chapter
  of the Association for Computational Linguistics}, pages 1--15, Dublin,
  Ireland.

\bibitem[\protect\citename{Trujillo}1994]{trujillo94}
Trujillo, Arturo.
\newblock 1994.
\newblock Computing {FIRST} and {FOLLOW} functions for {Feature-Theoretic}
  grammars.
\newblock In {\em COLING-94 - The 15th International Conference on
  Computational Linguistics}, pages 875--80, Kyoto, Japan.

\bibitem[\protect\citename{Trujillo and Berry}1996]{trujilloetal96}
Trujillo, Arturo and Simon Berry.
\newblock 1996.
\newblock Connectivity in bag generation.
\newblock In {\em COLING-96 - The 16th International Conference in
  Computational Linguistics}, pages 101--106, Copenhagen,
  Denmark.

\bibitem[\protect\citename{Whitelock}1994]{whitelock94}
Whitelock, Pete.
\newblock 1994.
\newblock Shake-and-bake translation.
\newblock In Rupp, C.~J., M.~A. Rosner, and R.~L. Johnson, editors.
\newblock {\em Constraints, Language and Computation}.
\newblock Academic Press, London, pages 339--59.

\end{thebibliography}

\end{document}